%% file: neurips_2026.tex
\documentclass{article}

\usepackage[preprint]{neurips_2026}

\usepackage[utf8]{inputenc} % allow utf-8 input
\usepackage[T1]{fontenc}    % use 8-bit T1 fonts
\usepackage{hyperref}       % hyperlinks
\usepackage{url}            % simple URL typesetting
\usepackage{booktabs}       % professional-quality tables
\usepackage{amsfonts}       % blackboard math symbols
\usepackage{nicefrac}       % compact symbols for 1/2, etc.
\usepackage{microtype}      % microtypography
\usepackage{xcolor}         % colors
\usepackage{natbib}
\usepackage{enumitem}
\usepackage{graphicx}
\usepackage{subcaption}
\usepackage{amsmath}
\usepackage{svg}
\svgsetup{inkscapelatex=false}
\workshoptitle{Representations for the Physical Sciences Workshop}
\title{Learning Hierarchical Causal Representations of the Effects of Forcings on Temperature in Climate Models}

\author{
\textbf{Shan Zhao}$^{1,2}$ \quad
\textbf{Ilija Trajkovic}$^{3}$ \quad
\textbf{Julia Kaltenborn}$^{2,4}$ \quad
\textbf{Yaniv Gurwicz}$^{5}$ \\
\textbf{Peer Nowack}$^{3}$ \quad
\textbf{David Rolnick}$^{2,4}$ \quad
\textbf{Julien Boussard}$^{2,4}$ \\[0.8em]
$^{1}$Technical University of Munich, Munich, Germany \\
$^{2}$Mila -- Quebec AI Institute, Montreal, Canada \\
$^{3}$Karlsruhe Institute of Technology, Karlsruhe, Germany \\
$^{4}$McGill University, Montreal, Canada \\
$^{5}$Intel Labs, Israel
}

\begin{document}

\maketitle

\begin{abstract}
%JK suggestion for the abstract (somewhat along those lines):
Machine learning (ML) emulators provide a fast and cost-effective method to simulate climate change scenarios after being trained on Earth System Models projections. However, the black-box nature of those data-driven approaches limit the usability and trustworthiness of their outputs and in particular their use as causal attribution tools. Here, we develop a hierarchical causal representation learning framework applied to sea surface temperature fields from a state-of-the-art global climate model. As a key advance over previous work, our framework explicitly models both atmospheric dynamical interactions arising from internal climate variability and forced responses due to changes in atmospheric greenhouse gas and aerosol concentrations. 
When trained on future climate change scenarios, our method accurately predicts the long-term global mean and regional temperature evolution and shows physically realistic responses to perturbations in greenhouse gas and aerosol concentrations when evaluated on unseen scenarios. Our results underline the potential of causal representation learning frameworks for advancing climate model emulation. 
% , for example by offering the possiblity to interact with the causal representation of a learned climate model. We extend on this work and develope a hierarchical causal representation learning framework that separates internal from forced variability. We present first results on disentangling the effects of local and global forcing concentrations on sea-surface temperature in NorESM2-LM.The model is trained on global NorESM2-LM forcings concentrations under different future climate change scenarios [shared socio-economic pathways] and evaluated on unseen climate projections. Our method accurately predicts long-term global mean temperature and shows physically realistic responses to forcing perturbations. 

% Data-driven approaches have revolutionized weather forecasting but struggle at emulating long-term climate variability. Causal discovery approaches have been successful at modelling climate teleconnections, but are limited to stationary scenarios and 

% We evaluate our method quantitatively on a semi-real dataset where the true causal graph is known, and qualitatively on the CMIP6 model projections. 
\end{abstract}

\section{Introduction}

Traditional climate models are based on a large set of differential equations that represent an enormous variety of processes and interactions driving the evolution of Earth's climate. Solving these equations over long time horizons is highly computationally expensive and requires major approximations, including limited numerical grid resolutions and various parameterizations for processes such as cloud formation and convection
% in the parametrization of the equations 
\citep{balaji_cpmip_2017, mansfield_predicting_2020}. This has led to a rapidly growing interest in ML approaches to climate modelling, given fast ML inference and the possibility to train the resulting models not only on climate model data, but also directly on Earth observations~\citep{wattmeyer2023acefastskillfullearned, clyne2026archesclimateprobabilisticdecadalensemble}.

% Machine-learning (ML) models can potentially emulate complex climate models and perform rapid simulations at inference.  
While ML models can now outperform traditional, physics-based models at medium-range weather forecasting in several metrics \citep{bi_accurate_2023, lam_graphcast_2023, weather_generator, bodnar_foundation_2024}, many of them have been demonstrated to show physically unrealistic behaviour 
% and instabilities 
when applied to long-term climate prediction \citep{chattopadhyay_challenges_2024, karlbauer_advancing_2024}. Few ML models can address the task of climate projection -- predicting the evolution of future climate given a scenario of greenhouse gas (GHG) and aerosol emissions -- while also realistically simulating short-to-long-term weather and climate variability. The international community urgently needs the ability to project future climate under defined shared socioeconomic pathway (SSP) scenarios \citep{RIAHI2017153}, which provide different estimates of future emissions of anthropogenic forcings (GHG and aerosols). 

The task of climate projection is extremely challenging given that anthropogenic climate change represents a long-term distribution shift in the presence of substantial natural internal variability across a large variety of spatial and temporal scales. Future observations come from a different distribution than past observations, while, ideally, a model should also be able to realistically capture short-term internal weather (hours-to weeks) to climate (years/decades) variability. ML models, which famously struggle with out-of-distribution (OOD) predictions~\citep{ovadia2019can}, cannot be trusted in this context. Simple models, e.g.~linear pattern scaling, cannot capture the complex processes underlying the evolution of the climate system, while increasing model complexity does not necessarily improve the emulation of the climate response to forcing scenarios, as expressive models may overfit correlations specific to training scenarios \citep{bjorn_linear}. 
Causal representation learning has been proposed as a way to automatically learn a physically meaningful representation of the climate system, by 
learning the causal teleconnections between the observed variables of the modeled system \citep{brouillard_causal_2024, boussard2023towards, picabu}. The learned causal graph can thus represent both internal dynamics e.g. how the climatic variables naturally interact, and the response to external forcings. These score-based methods leverage differentiable constraints \citep{zheng_dags_2018, brouillard_differentiable_2020} and have been applied to emulate the dynamics of sea-surface temperature in a pre-industrial control scenario e.g. without external forcings (PICABU, \citealp{picabu}), succesfully modeling the internal climate variability. 
% Causal ML proposes to overcome these challenges by automatically discovering the causal teleconnections between the observed variables of the modeled system. In particular, PCMCI \citep{runge_causal_2018, runge_inferring_2019, gerhardus_high-recall_2020} has been applied to the climate system and can be combined with dimensionality reduction methods to obtain low-dimensional latent variables before learning causal connections between these latents \citep{nowack_causal_2020, tibau_spatiotemporal_2022, falasca_data-driven_2024}. Causal representation learning extends on these approaches to jointly learn a causal latent representation from high-dimensional data. Fully differentiable approaches have been developed \citep{zheng_dags_2018, brouillard_differentiable_2020}, and applied to emulate sea-surface temperature \citep{brouillard_causal_2024, boussard2023towards, picabu}. 
% Building upon PICABU \citep{picabu}, a causal representation learning framework developed to automatically learn the teleconnections underlying the climate system to represent the internal climate variability. 
In this paper, we learn to model the effect of various anthropogenic forcings on temperature, and explicitly separate the internal from the externally-driven variability of the climate system, at different spatial scales. We show that our model captures scenario-dependent warming trends and produces distinct temperature responses to GHG and regional aerosol perturbations.  
% Storyline: simple models outperform complex AI emulators on predicting temperature under different scenarios. We build an interpretable causal method to disentangle internal from externally driven variability, and the effects of the different local and global greenhouse gases. Natural variability can sometimes exceed externally-driven variability and hinder attribution studies e.g. the attribution of extreme events to anthropogenic emissions. 
This work is a first step towards principled, causal, learned representations of the climate system, and the design of statistical attribution studies. 

% Our contributions are as follows: 
% \begin{itemize}[nosep]
%     \item We extend existing approaches to model the effect of the different anthropogenic forcings on temperature, explicitly separating the internal from the externally-driven variability of the climate system, at different spatial scales. 
%     \item We extend a dataset, developed for emulating sea-level pressure with known underlying connections, to incorporate multiple variables and forcings and evaluate our model and baselines on this simulated dataset. 
%     \item We show that our method is able to separate internal from externally-driven variability of the climate system. 
% \end{itemize}

% \subsection{Related work}

% ML methods are being increasingly used for Earth System Sciences, in particular for medium-range weather forecasting \textbf{cite} where they have outperformed traditional models. Approaches have been proposed for climate modelling but physical realism and out-of-distribution generalization remain unsolved challenges. Approaches include long-term autoregressive predictions \textbf{cite}, physics-informed neural networks or constrained predictions for improved physical consistency, and foundation models. 

\input{Chapters/methods}

\input{Chapters/results}

\section{Discussion}

Our hierarchical causal representation learning model is able to accurately predict the GMST from GHG and aerosol concentrations in one moderate extrapolation scenario. Furthermore, it allows us to disentangle local from global effects, as we explicitly condition local and global representations on forcings of different spatial scales. It is also able to distinguish the effects of two different aerosol concentrations, BC and SO$_2$, over East Asia. However, the current formulation does not guarantee separation of individual forcings within the same hierarchical level, such as CO$_2$ and CH$_4$ at the global level. Future work will incorporate causal modules to better identify the effects of individual forcings within each level \citep{wang2019blessingsmultiplecauses, almodovar2025decaflowdeconfoundingcausalgenerative}. 

The evaluation of our model is limited to realistic representations of the forced response and ENSO power spectra within a climate model, as there is no ground truth causal graph of, e.g. atmospheric dynamical interactions. Although there exist benchmarks for evaluating causal discovery \citep{herdeanu2025causaldynamicslargescalebenchmarkstructural} and causal representation learning (SAVAR, \citealp{tibau_spatiotemporal_2022}), none of these benchmarks includes external forcings. In future work, we will extend SAVAR with external forcings to evaluate our model in a controlled setting, and benchmark it against a set of existing methods. 
Moreover, we will include more climatic variables, such as sea level pressure or precipitation, to give a more complete picture of the causal connections between forcings and climate variables. We will also evaluate our method on additional climate models (e.g. CESM2) and more challenging scenarios (e.g. 4$\times$CO$_2$, SSP5-8.5). Finally, we hope to apply this work to observed extreme events, and perform a proper attribution study in a real-world use case. 

\section{Acknowledgements}
This project was supported by the Schmidt Sciences AI2050 program, the Sloan Fellowship program, IVADO, Canada CIFAR AI Chairs program, and Canada First Research Excellence Fund. We acknowledge computational support from Mila – Quebec AI Institute, including in-kind support from Nvidia Corporation. P.N. was supported by the WOW – a World model of Our World project, funded by the Carl-Zeiss-Stiftung.

\newpage
% \section*{Rfeferences}

\bibliography{refs}
\bibliographystyle{plainnat}

%%%%%%%%%%%%%%%%%%%%%%%%%%%%%%%%%%%%%%%%%%%%%%%%%%%%%%%%%%%%
\newpage
\clearpage
\appendix
\input{Chapters/appendix}

% \section{Technical appendices and supplementary material}
% Technical appendices with additional results, figures, graphs, and proofs may be
% included in the same PDF as the main paper. There is no page limit for the
% technical appendices. Additional files are not allowed, but authors may link to
% properly anonymized code and/or data repositories.

% Note: Think of the appendix as ``optional reading'' for reviewers. The paper must be able to stand alone without the appendix; for example, adding critical experiments that support the main claims to an appendix is inappropriate. 

%%%%%%%%%%%%%%%%%%%%%%%%%%%%%%%%%%%%%%%%%%%%%%%%%%%%%%%%%%%%

\end{document}

%% file: Chapters/methods.tex
\section{Methods} \label{sec:methods}
\begin{figure}
    \centering
    \includegraphics[width=\linewidth,trim={0cm 9cm 3.4cm 0cm},clip]{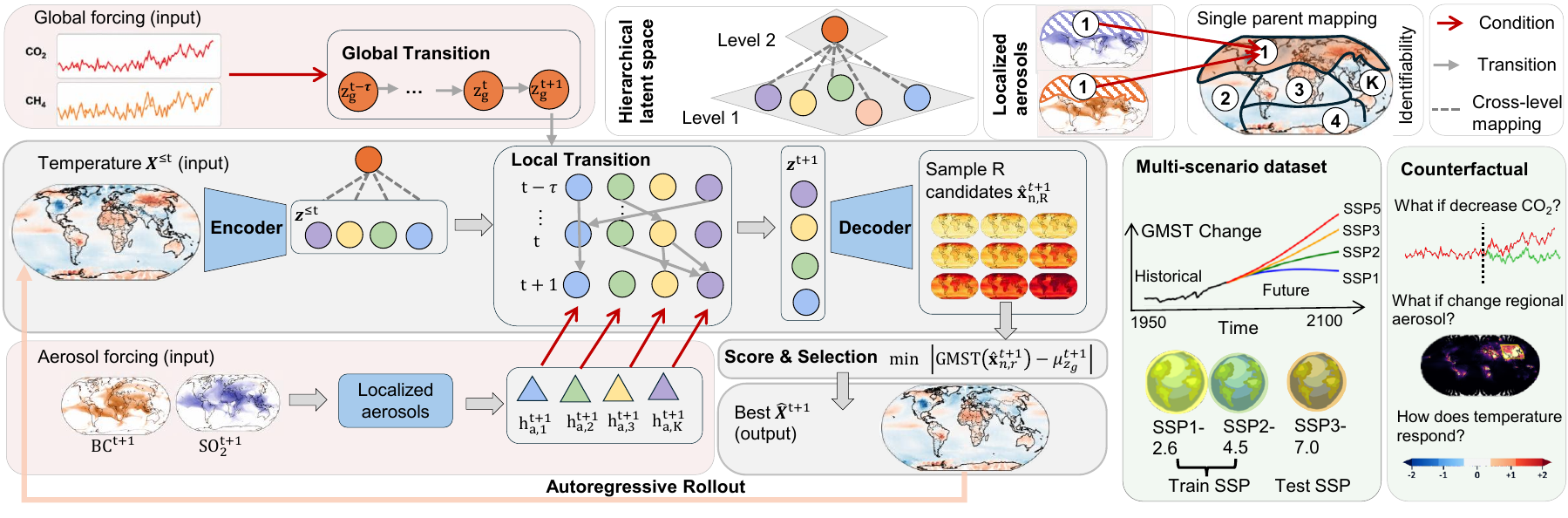}
    \caption{\small \textbf{Hierarchical climate emulator.}
A global latent $z_g$ captures climate responses to global forcings,
while local latents $\mathbf{z}$ capture regional dynamics conditioned on
spatially localized aerosols under the single-parent mapping. The model supports multi-scenario emulation and counterfactual perturbations of forcings.}
    \label{fig:model_full}
\end{figure}

\subsection{Model}

PICABU~\citep{picabu} jointly learns a low-dimensional latent representation $\{\mathbf{z}^t=(z_1,...,z_K)^t\}_{t \in \{t-\tau, ..., t-1, t\}}$ of the high-dimensional climate states $\{\mathbf{x}\}_{t \in \{t-\tau, ..., t-1, t\}}$ and their dynamics, represented as sparse causal connections from $\mathbf{z}^{<t}$ to $\mathbf{z}^t$. PICABU adopts the single-parent assumption, i.e., at a given timestep, each spatial grid point is mapped to a single latent $z_k$, resulting in $K$ spatial clusters of grid points sharing similar climate patterns. This constrains the latent-to-climate mapping $\mathbf{W}$ to satisfy the structural conditions required for identifiability. The model is trained by maximizing the evidence lower bound (ELBO) to maximize the probability of climate states (both in reconstructing $x^{<t}$ and predicting $x^t$) under constraint $\mathcal{C}_{\mathrm{single~parent}}^{\{\lambda,\mu\}}$, using the augmented Lagrangian method (ALM). Additionally, the temporal dependencies between local latents are constrained by $\mathcal{C}_{\mathrm{sparsity}}^{\{\lambda,\mu\}}$ to form a sparse transition graph. %Authors additionally add a term $\lambda_a\mathcal{L}_{\mathrm{aux}}$ to the loss to optimize for the CRPS and spatio-temporal spectrum of the predicted temperature. The core objective is 
% \begin{equation}
% \small
% \mathcal{L}_{\mathrm{PICABU}}^{\{\lambda,\mu\}} = -\mathcal{L}_{\mathrm{ELBO}} + \lambda_a\mathcal{L}_{\mathrm{aux}} +\mathcal{C}_{\mathrm{single~parent}}^{\{\lambda,\mu\}} + \mathcal{C}_{\mathrm{sparsity}}^{\{\lambda,\mu\}}.
% \label{eq:core_loss_picabu}
% \end{equation}
% -------------------------------------------------------------- 
\paragraph{Hierarchical latent representation.} 
Building on this structure, we introduce a hierarchical latent representation and incorporate external climate forcings at different levels of the hierarchy. At the highest level, we introduce a global latent $z_g\in\mathbb{R}$, which yields the hierarchy $z_g\rightarrow\mathbf{z}\rightarrow\mathbf{x}$ that retains the single-parent structure across both latent levels and thus preserves identifiability. Temporal dynamics are modeled at both levels:

\begin{equation}
\small
\underbrace{(z_g^{<t},\,\mathrm{CO}_2^t,\,\mathrm{CH}_4^t)\rightarrow z_g^t}_{\text{global transition}},\qquad
\underbrace{(z_g^t,\,\mathbf{z}^{<t},\,\mathrm{BC}_k^t,\,\mathrm{SO}_{2,k}^t)\rightarrow z_k^t}_{\text{local transition}},
\label{eq:hierarchical_transition}
\end{equation}
where global GHG forcings $\mathrm{CO}_2,\,\mathrm{CH}_4$ (carbon dioxide, methane) act through $z_g$, while aerosol forcings affect local latents. $\mathrm{SO}_{2,k}$ and $\mathrm{BC}_k$ denote the localized aerosol forcings (sulfur, black carbon) associated with $z_k$, as detailed in Eq.~\ref{eq:aerosol}. 
% -------------------------------------------------------------- 
% \subsubsection{Localized forcing} \label{method:local_forcing}
To capture localized aerosol effects, we use $\mathbf{W}$ to map aerosols into the same $K$ clusters, and each climate latent is affected by aerosols only within its corresponding region. For $a\in\{\mathrm{BC},\mathrm{SO}_2\}$, the localized forcing used to condition $z_k$ is transformed by the nonlinear function $f_{a,k}$:
\begin{equation}
\small
\mathbf{h}_{a,k}^{t} = f_{a,k}\!\left(\mathbf{W}_{:,k}\odot\mathbf{a}^{t}\right).
\label{eq:aerosol}
\end{equation}
Therefore, $\mathbf{W}$ jointly represents the spatial mapping to capture both climate variability and regional responses to aerosol forcing. We adapt the ELBO to the hierarchical model, now called $\mathcal{L}_{\mathrm{H-ELBO}}$ by factorizing the posterior and prior according to the latent hierarchy, decomposing the standard KL term into local- and global-level KL divergences. Details are provided in~\ref{app:methods}.
% \begin{equation}
% \small
% \mathcal{L}_{\mathrm{core}}^{\{\lambda,\mu\}} = -\mathcal{L}_{\mathrm{H-ELBO}} + \lambda_a\mathcal{L}_{\mathrm{aux}} + \mathcal{C}_{\mathrm{single~parent}}^{\{\lambda,\mu\}} + \mathcal{C}_{\mathrm{sparsity}}^{\{\lambda,\mu\}}.
% \label{eq:core_loss}
% \end{equation}

\paragraph{Global temperature constraints.}
We introduce constraints on global mean surface temperature (GMST) at both the latent and observation levels. We encourage both the global latent and the global mean of the predicted temperature field to be close to the true GMST during training. With $\mu_x^t$  the spatial mean of the predictive distribution $p(x^t\mid\mathbf{z}^t)$ and $\mu_g^t$ the mean of $p(z_g^t\mid z_g^{<t})$, we add the following loss function to our training objective: 
\begin{equation}
\small
\mathcal{L}_{\mathrm{GMST}}
= \left\|\mu_g^t-\mathrm{GMST}^t\right\|_2^2
% , \qquad \mathcal{L}_{x_g} = 
+ \left\|\mu_{x}^t-\mathrm{GMST}^t\right\|_2^2.
\label{eq:gmst_losses}
\end{equation}
% While $\mathcal{L}_{z_g}$ provides the global latent with an explicit physical interpretation, $\mathcal{L}_{x_g}$ constrain the global mean of the predicted temperature field while allowing for redistribution of temperature anomalies across regions.

The overall core training objective is in Eq.\ref{eq:loss}, where $\mathcal{L}_{\mathrm{aux}}$ comprises the probabilistic and spectral losses in PICABU \citep{picabu}:
\begin{equation}
\small
\mathcal{L}_{\mathrm{core}} := -\mathcal{L}_{\mathrm{H-ELBO}} +\mathcal{C}_{\mathrm{single~parent}}^{\{\lambda,\mu\}} + \mathcal{C}_{\mathrm{sparsity}}^{\{\lambda,\mu\}}+\lambda_g\mathcal{L}_{\mathrm{GMST}} + \lambda_a\mathcal{L}_{\mathrm{aux}}
\label{eq:loss}
\end{equation}
% where $\mathcal{L}_{\mathrm{aux}}$ comprises the CRPS and temporal and spatial
% spectral losses in ~\cite{picabu}. 

\subsection{Hierarchical autoregressive rollout}\label{sec:method_rollouts}
To generate long-term climate trajectories, we perform autoregressive rollouts by first predicting the global latent value at time $T+1$ given the global forcing concentrations at $T+1$, and then predicting the local latents given the global latent value and local forcing concentrations, before decoding the temperature fields e.g. going from top to bottom in the hierarchy. We iterate this procedure to get full trajectories. 
% At each timestep, the predicted temperature field from the previous steps is used as the current climate state. 
Since our model is probabilistic, we generate $N > 1$ trajectories. We initially draw $N$ samples for the local latents, and, at each timestep, for each sample, we sample $R$ candidates $\mathbf{z}_{n,r}^{t+1}\sim p(\mathbf{z}^{t+1}\mid\mathbf{z}_n^{\leq t},z_g^{t+1})$ before keeping the highest-scoring candidate, yielding $N$ filtered trajectories for autoregressive propagation. 
The score associated with each sample is the L2 norm between the mean of the global latent $\mu_g^{t+1}$ and the mean of the decoded temperatures. Thus, we ensure that the values of the local latents lead to a global mean temperature that is close to the global latent value, which itself is trained to approximate the true GMST. This is an extension of the Bayesian filtering technique used in PICABU, where we use our own predictions of the GMST rather than assuming a fixed spatial spectrum of the temperature field. 

% The second, denoted by $\mathcal{S}_g$ and introduced for global-forcing perturbation experiments, scores each candidate based on the consistency between the GMST of its decoded prediction and the predicted $\mu_g^{t+1}$. This explicitly incorporates the forcing-driven global response into particle selection, allowing the perturbed global forcing signal encoded in $z_g$ to propagate to the decoded climate trajectories.

% We consider two strategies for scoring candidate particles. The first, denoted by $\mathcal{S}_l$, follows PICABU and scores each sampled local latent by its log-probability under the predicted transition distribution. 

%% file: Chapters/results.tex
\section{Experiments and results}
\subsection{Experiments}
We train and evaluate our model on monthly data from NorESM2-LM~\citep{seland2020overview} simulations. 
It is trained to predict the monthly surface temperature from previous months' surface temperatures and CO$_2$, CH$_4$, BC, and SO$_2$ concentrations at the current month, as given by the SSP scenarios. 
We train on the historical period, SSP1-2.6, and SSP2-4.5, and test on the OOD SSP3-7.0 scenario. 
CO$_2$ and CH$_4$ are included as global GHG forcings and BC and SO$_2$ as spatially varying aerosol forcings. Temperature is deseasonalized using the pre-industrial monthly climatology, without forced trend, while all variables are standardized using historical statistics. We use $K=48$ local latents, one global latent, $d_x=3072$ grid points on a HEALPix grid (equal area grid, \citep{Gorski_2005}), and $\tau=5$ months. Rollouts use $N=50$ particles with $R=10$ candidates each. Full details are in~\ref{app:parameters}.

\subsection{Results}
\begin{figure}
    \centering
    \includegraphics[width=\linewidth,trim={0cm 10.6cm 2cm 0cm},
            clip
        ]{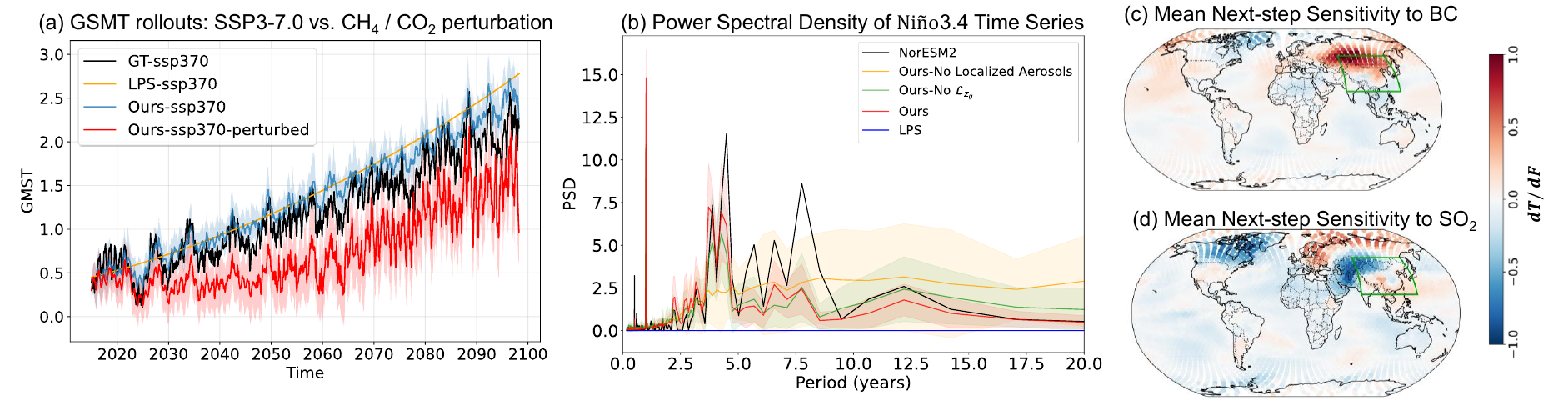}
\caption{\small \textbf{Long-term climate rollouts and forcing perturbations.}
\textbf{(a)} Our GMST rollouts under SSP3-7.0 (blue) compared with NorESM2 (black) and Linear Pattern Scaling (LPS) prediction (orange). Fixing CO$_2$ and CH$_4$ at their 2015 levels (red) suppresses the long-term increase in GMST.
\textbf{(b)} Power spectral density of the Ni\~no3.4 index under SSP3-7.0. The model ensemble mean (red) reproduces the characteristic Ni\~no3.4 spectral peak. Shading in (a,b) denotes the std. across 50 trajectories. Details of the ablations and comparison to LPS are in \ref{app:ablation}.
\textbf{(c,d)} Next-step temperature responses to regional BC and SO$_2$ perturbations over East Asia, showing warming and cooling responses, respectively. Green boxes indicate the perturbed region.}
    \label{fig:result}
\end{figure}

We evaluate long-term climate emulation using 85-year autoregressive rollouts from January 2015.
% with scoring strategy $\mathcal{S}_l$ (Section~\ref{sec:method_rollouts}). 
Despite being trained only on SSP1-2.6 and SSP2-4.5, the model captures the stronger warming under the held-out SSP3-7.0 scenario (Fig.~\ref{fig:result}a). While still overestimating the overall warming trend, ours achieves the closer mean to NorESM2 for both GMST ($1.41$ vs. $1.24$) and Ni\~no3.4 ($1.47$ vs. $1.37$) compared with baseline. The model also captures the characteristic Ni\~no3.4 periodicity of approximately 2.5--5 years and the annual peak corresponding to seasonal variations (Fig.~\ref{fig:result}b). Additional rollout analyses are provided in \ref{app:main_results}.

We then evaluate temperature responses to forcing perturbations. Fixing CO$_2$ and CH$_4$ at their 2015 levels suppresses the long-term increase in $z_g$ (~Fig.~\ref{fig:rollout_pz2mu}), accompanied by reduced GMST warming (Fig.~\ref{fig:result}a). The perturbed trajectories also exhibit a larger ensemble spread, indicating increased uncertainty under the modified forcing conditions. 
% Candidate selection uses $\mathcal{S}_g$ to propagate the perturbed global-forcing signal through $z_g$.
Regional aerosol perturbations produce localized, forcing-specific responses (Fig.~\ref{fig:result}c,d): increasing BC over East Asia induces temperature increases, whereas increasing SO$_2$ produces temperature decreases, consistent with their dominant warming and cooling radiative effects, respectively~\cite{ramanathan2009air}. Additional forcing--response analyses are provided in \ref{app:counterfactual}.

% \begin{figure}[htbp!]
%     \centering
%     \includegraphics[width=1\textwidth,trim={0cm 13cm 14.5cm 0cm},clip]{Figures/perturbation_global_forcing_rollouts.pdf}
%     \caption{\textbf{85-year rollouts under perturbed global forcings (SSP3-7.0).}
% We evaluate the model under a counterfactual SSP3-7.0 scenario in which both CO$_2$ and CH$_4$ are held fixed at their 2015 levels throughout the 85-year autoregressive rollout. Holding CO$_2$ and CH$_4$ fixed substantially suppresses the long-term increase in the global climate latent, with a corresponding reduction in the GMST warming. }
%     \label{fig:perturb_global}
% \end{figure}

% \begin{figure}[htbp!]
%     \centering
%     \includegraphics[width=.75\textwidth,trim={0cm 15cm 23cm 0cm},clip]{Figures/perturbation_local_forcing_sensativity.pdf}
% \caption{\textbf{ Mean next-step temperature sensitivity to regional SO$_2$ (left) and BC (right) perturbations} over East Asia  ($73^\circ$E--$135^\circ$E, $18^\circ$N--$54^\circ$N). For each forcing, additive perturbations spanning 50 levels from $-1.5$ to $1.5$ are applied within the selected region while the remaining forcing fields are kept unchanged. The sensitivity $dT/dF$ is estimated as the linear slope of the predicted temperature response with respect to the forcing perturbation at each spatial location, separately for 30 randomly selected initial conditions, and subsequently averaged across these samples. The yellow boxes indicate the perturbed region.  }
%     \label{fig:aerosol_pert_sensitivity}
% \end{figure} 

%% file: Chapters/appendix.tex
\section{Methods}\label{app:methods}
\subsection{Hierarchical Generative Model}\label{app:model}
\begin{figure}
    \centering
    \includegraphics[width=.75\linewidth,trim={0cm 3.2cm 2.2cm 0cm},clip]{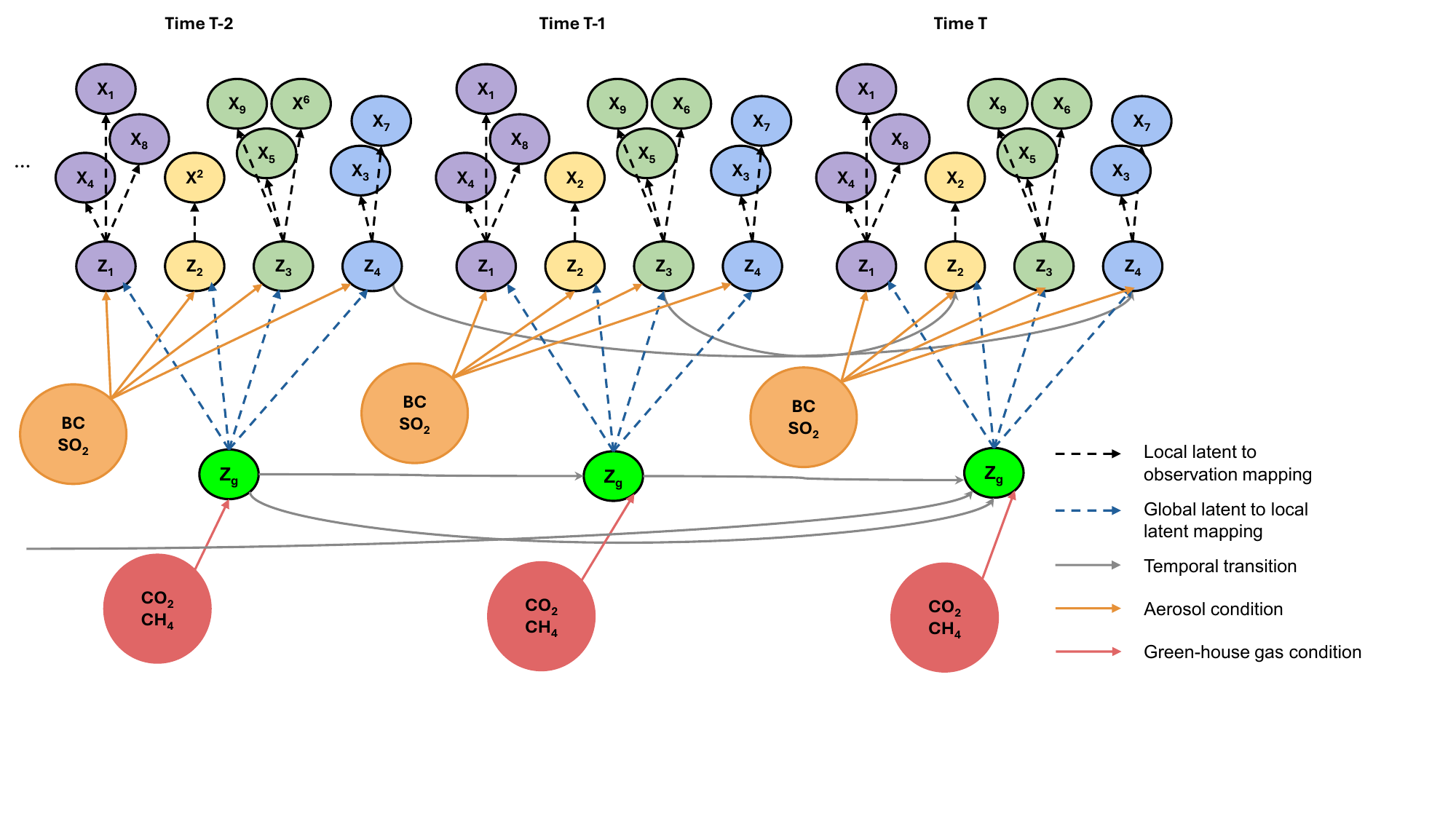}
    \caption{\textbf{Overview of the hierarchical generative model.}
    At each timestep, greenhouse-gas forcings (CO$_2$ and CH$_4$) condition the
    global latent $z_g$, whose temporal evolution depends on its previous states.
    The resulting global state conditions the local latents $\mathbf{z}$, which
    additionally depend on their sparse temporal parents and spatially localized
    aerosol forcings (BC and SO$_2$). Aerosol forcing is assigned to each local
    latent according to the learned latent-to-observation mapping, while each
    observation is generated from its associated local latent under the
    single-parent structure. This hierarchy enables global greenhouse-gas
    responses and regional aerosol effects to enter the climate dynamics through
    different spatial levels.}
    \label{fig:model_simple}
\end{figure}

Let $\mathbf{x}^t\in\mathbb{R}^{d_x}$ denote the spatial temperature field at time $t$. We represent the climate state by $K$ local latent variables $\mathbf{z}^t=(z_1^t,\ldots,z_K^t)$ and a global climate latent $z_g^t$. The local climate representation is inferred from the temperature field, followed by the global representation, 
\begin{equation} 
q(\mathbf{z}^t,z_g^t\mid\mathbf{x}^t) = q_\phi(\mathbf{z}^t\mid\mathbf{x}^t) q_\psi(z_g^t\mid\mathbf{z}^t). 
\label{eq:posterior} 
\end{equation} 
We model the global latent dynamics conditioned on greenhouse-gas forcings as 
\begin{equation} 
p\!\left( z_g^t \mid z_g^{<t}, \mathrm{CO}_2^t, \mathrm{CH}_4^t \right), 
\label{eq:global_dynamics} 
\end{equation} 
whose parameters are given by 
\begin{equation} 
(\mu_g^t,\sigma_g^t) = f_g\!\left( z_g^{<t}, \mathrm{CO}_2^t, \mathrm{CH}_4^t \right). 
\end{equation} 
Conditioned on the current global state, each local climate latent follows 
\begin{equation} 
p\!\left( z_k^t \mid \mathrm{Pa}(z_k^t), z_g^t, \mathbf{a}_{k}^t \right), 
\label{eq:local_dynamics} 
\end{equation} 
where $\mathrm{Pa}(z_k^t)\subseteq\mathbf{z}^{<t}$ is determined by the transition graph under $\mathcal{C}_{\mathrm{sparsity}}^{\{\lambda,\mu\}}$ and $\mathbf{a}_{k}^t$ denotes the localized aerosol forcing associated with latent $k$. 

The local transition is then parameterized as
\begin{equation}
(\mu_k^t,\sigma_k^t) = g_k\!\left(z_g^t, \mathrm{Pa}(z_k^t), \mathbf{h}_{\mathrm{BC},k}^t,\mathbf{h}_{\mathrm{SO}_2,k}^t\right),
\label{eq:local_transition}
\end{equation}
where $\mathbf{h}_{\mathrm{BC},k}^t$ and
$\mathbf{h}_{\mathrm{SO}_2,k}^t$ denote the localized BC and SO$_2$
forcing representations in Eq.~\ref{eq:aerosol}. The overall graphical model is illustrated in Fig.~\ref{fig:model_simple}.

\subsection{Hierarchical ELBO Derivation}
\label{app:helbo}

We extend the original variational objective to a two-level latent hierarchy. Starting from the sequential variational objective,
\begin{equation}
\begin{aligned}
\log p(\mathbf{x}^{\leq T}) \geq \sum_{t=1}^{T} \Big[ &\mathbb{E}_{q(\mathbf{z}^t\mid\mathbf{x}^t)}
    \left[\log p(\mathbf{x}^t\mid\mathbf{z}^t)\right]-\mathbb{E}_{q(\mathbf{z}^{<t}\mid\mathbf{x}^{<t})}\mathrm{KL}\!\left[q(\mathbf{z}^t\mid\mathbf{x}^t)\,\Vert\,p(\mathbf{z}^t\mid\mathbf{z}^{<t})\right]\Big].
\end{aligned}
\label{eq:elbo_climate}
\end{equation}

Let $\mathbf{z}_1^t$ denote the local climate latents and $z_2^t$ the high-level global latent. 
The hierarchical posterior is factorized as
\begin{equation}
q(\mathbf{z}_1^t,z_2^t\mid\mathbf{x}^t)= q(\mathbf{z}_1^t\mid\mathbf{x}^t) q(z_2^t\mid\mathbf{z}_1^t,\mathbf{x}^t) = q(\mathbf{z}_1^t\mid\mathbf{x}^t) q(z_2^t\mid\mathbf{z}_1^t). \label{eq:hierarchical_posterior}
\end{equation}
where the last equality follows from the dependency structure encoded in our graphical model, in which $z_2^t$ is conditionally independent of $\mathbf{x}^t$ given $\mathbf{z}_1^t$.

The reconstruction term (first term) becomes
\begin{align}
\mathbb{E}_{q(z_2^t\mid\mathbf{z}_1^t)q(\mathbf{z}_1^t\mid\mathbf{x}^t)}\left[\log p(\mathbf{x}^t\mid\mathbf{z}_1^t)\right]=\mathbb{E}_{q(\mathbf{z}_1^t\mid\mathbf{x}^t)}\left[\log p(\mathbf{x}^t\mid\mathbf{z}_1^t)\right],
\label{eq:hierarchical_reconstruction}
\end{align}
where the observations are decoded directly from the local latent representation $\mathbf{z}_1^t$.

The prior follows the two-level hierarchy
\begin{equation}
p(\mathbf{z}_1^t,z_2^t\mid \mathbf{z}_1^{<t},z_2^{<t}) =p(z_2^t\mid \mathbf{z}_1^{<t},z_2^{<t})p(\mathbf{z}_1^t\mid \mathbf{z}_1^{<t},z_2^{<t},z_2^t)=
p(z_2^t\mid z_2^{<t})p(\mathbf{z}_1^t\mid \mathbf{z}_1^{<t},z_2^t),
\label{eq:hierarchical_prior}
\end{equation}

where the last equality follows from the conditional dependencies specified by our graphical model, 
\begin{equation}
z_2^t \perp\!\!\!\perp z_1^{<t}\mid z_2^{<t},
\qquad
z_1^t \perp\!\!\!\perp z_2^{<t}
\mid (z_1^{<t},z_2^t).
\end{equation}

Accordingly, the KL divergence term (second term) becomes 
\begin{align*}
\mathbb{E}_{q(\mathbf{z}_2^{<t} \mid \mathbf{z}_1^{<t})q(\mathbf{z}_1^{<t} \mid \mathbf{x}^{<t})} \text{KL}\left[q(\mathbf{z}_2^t \mid \mathbf{z}_1^t) q(\mathbf{z}_1^t \mid \mathbf{x}^t) \,||\, p(\mathbf{z}_2^t \mid \mathbf{z}_2^{< t})p(\mathbf{z}_1^t \mid \mathbf{z}_1^{< t}, z_2^t)\right] = \\\\ \mathbb{E}_{q(\mathbf{z}_2^{<t} \mid \mathbf{z}_1^{<t})q(\mathbf{z}_1^{<t} \mid \mathbf{x}^{<t})} (\text{KL}\left[q(\mathbf{z}_2^t \mid \mathbf{z}_1^t) \,||\, p(\mathbf{z}_2^t \mid \mathbf{z}_2^{< t})\right] + \text{KL}\left[q(\mathbf{z}_1^t \mid \mathbf{x}^t) \,||\, p(\mathbf{z}_1^t \mid \mathbf{z}_{1}^{< t}, z_2^t\right] ) \\\\
\end{align*}

Combining the reconstruction and the KL terms gives
\begin{align}
\mathcal{L}_{\mathrm{H\text{-}ELBO}}=\sum_{t=1}^{T}\Bigg[&\mathbb{E}_{q(\mathbf{z}_1^t\mid\mathbf{x}^t)}\left[\log p(\mathbf{x}^t\mid\mathbf{z}_1^t)\right]-\mathrm{KL}\!\left[q(\mathbf{z}_1^t\mid\mathbf{x}^t)\,\Vert\,p(\mathbf{z}_1^t\mid\mathbf{z}_1^{<t}, z_2^t)\right]\nonumber\\
&-\mathrm{KL}\!\left[q(z_2^t\mid\mathbf{z}_1^t)\,\Vert\,p(z_2^t\mid z_2^{<t})\right]\Bigg].
\label{eq:hierarchical_elbo}
\end{align}
\section{Hyperparameters}\label{app:parameters}
Table~\ref{tab:model_hparams} summarizes the model architecture and
training hyperparameters used for the reported experiments.

\begin{table}[h]
\centering
\caption{Model and training hyperparameters used in the experiments. We denote an MLP with hidden dimension $a$ and $b$ layers as $\mathrm{MLP}(a,b)$, and use $\mathrm{Linear}(a,b)$ analogously for the
corresponding linear layer. The model is trained with RMSProp ($\mathrm{lr}=10^{-4}$, batch size $128$) for up to $10^5$ iterations. The auxiliary loss term incorporates the continuous ranked probability score (CRPS) and a spatio-temporal spectral loss on the predicted temperature.}
\label{tab:model_hparams}
\begin{tabular}{ll}
\toprule
\textbf{Parameter} & \textbf{Value} \\
\midrule
\multicolumn{2}{l}{\textit{Latent representation}} \\
Number of spatial grid cells $d_x$ & 3072 \\
Number of local latent variables $d_z$ & 48 \\
Number of global latent variables $d_{z_g}$ & 1 \\
Temporal context length $\tau$ & 5 months \\
\midrule
\multicolumn{2}{l}{\textit{Encoder and decoder}} \\
Encoder& MLP(8,2)\\
Decoder & MLP(8,2) \\
Activation function & LeakyReLU \\
Encoder distribution & Gaussian \\
Decoder distribution & Gaussian \\
\midrule
\multicolumn{2}{l}{\textit{Latent dynamics}} \\
Local transition dynamics & MLP(8,2) \\
Global transition dynamics & Linear($d_z \times \tau$, 1) \\
Transition distribution & Gaussian \\
Global forcings & CO$_2$, CH$_4$ \\
Spatial forcings & BC, SO$_2$ \\
\midrule
\multicolumn{2}{l}{\textit{Loss coefficients}} \\
KL coefficient $\lambda_{\mathrm{KL}}$ & 1 \\
GMST coefficient $\lambda_{\mathrm{GMST}}$ & $10^3$ \\
CRPS coefficient $\lambda_{\mathrm{CRPS}}$ & 1 \\
Spatial spectral coefficient & $10^3$ \\
Temporal spectral coefficient & $10^3$ \\
\midrule
\multicolumn{2}{l}{\textit{Optimization}} \\
Optimizer & RMSProp \\
Initial learning rate & $10^{-4}$ \\
Batch size & 128 \\
Maximum training iterations & $10^5$ \\
Validation interval & 100 iterations \\
Early-stopping patience & 5000 iterations \\
\midrule
\multicolumn{2}{l}{\textit{Single-parent constraint (ALM)}} \\
Initial penalty $\mu_0$ & $10^4$ \\
Penalty multiplier & 1.2 \\
$\omega_\gamma$ & 0.01 \\
$\omega_\mu$ & 0.9 \\
Constraint threshold & $10^{-2}$ \\
Minimum convergence interval & 1000 \\
\midrule
\multicolumn{2}{l}{\textit{Transition sparsity constraint (ALM)}} \\
Initial penalty $\mu_0$ & $10^{-2}$ \\
Penalty multiplier & 1.2 \\
$\omega_\gamma$ & 0.01 \\
$\omega_\mu$ & 0.95 \\
Constraint threshold & $10^{-4}$ \\
Minimum convergence interval & 1000 \\
Graph binarization threshold & 0.5 \\
\midrule
\multicolumn{2}{l}{\textit{Rollout}} \\
Number of particles & 50 \\
Candidate samples per particle & 10 \\
\bottomrule
\end{tabular}
\end{table}

\section{Results} 
\subsection{Additional rollout analysis}\label{app:main_results}
\begin{figure}[htbp!]
    \centering
    \begin{subfigure}[t]{0.75\textwidth}
        \centering
        \includesvg[
            width=\linewidth,
            % trim={0cm 12.8cm 20cm 0cm},
            % clip
        ]{Figures/Figure_pz2mu/rollouts_pred_ssp126_ssp245_ssp370_pz2mu}
        \caption{GMST rollouts under SSP1-2.6, SSP2-4.5, SSP3-7.0.}
        \label{fig:rollouts_3ssp_temp}
    \end{subfigure}
    \hfill
    \begin{subfigure}[t]{0.95\textwidth}
        \centering
    \includegraphics[
        width=\linewidth,
        trim={0cm 5cm 0cm 0cm},
        clip
    ]{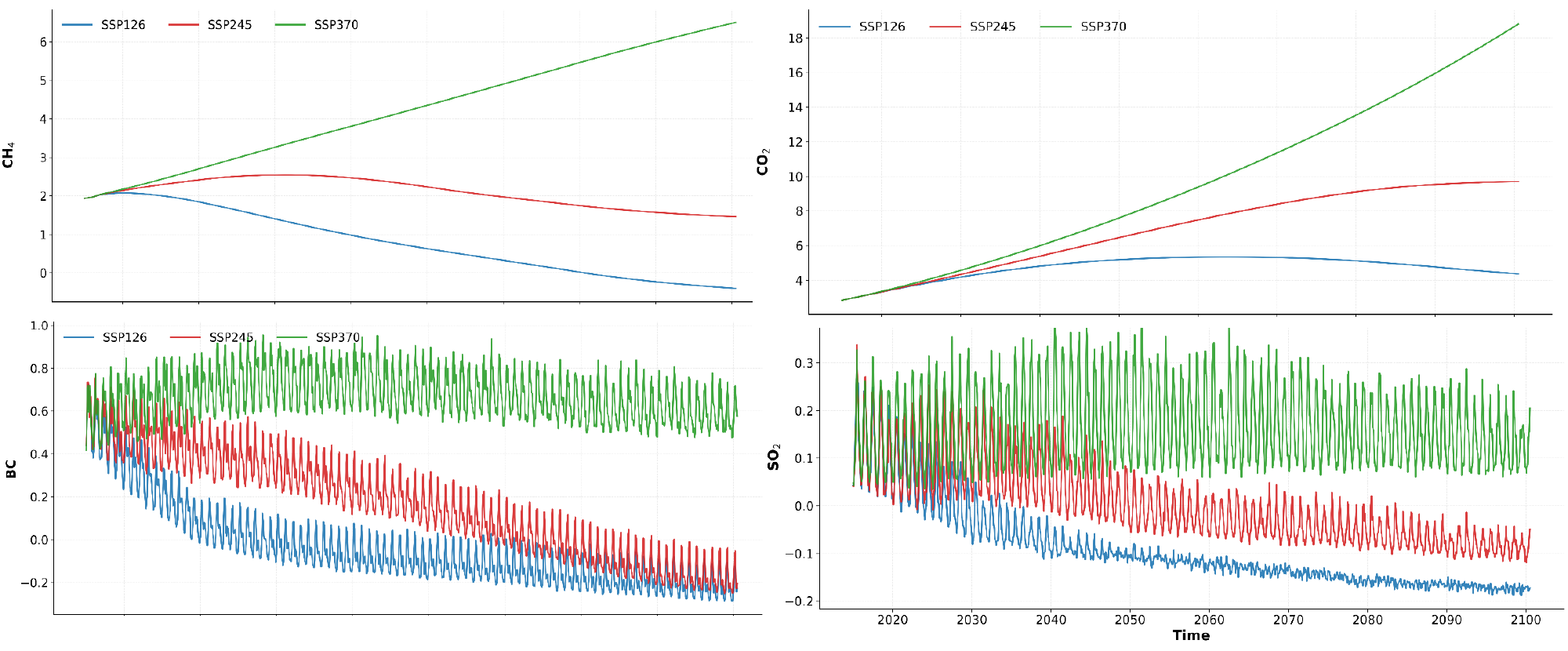}
    \caption{Preprocessed External forcings under the three SSP scenarios.}
        \label{fig:3ssp_forcings}
    \end{subfigure}

    \caption{
    \textbf{GMST rollouts and forcings.} 
    (a) GMST rollouts under SSP1-2.6 (blue), SSP2-4.5 (red), and SSP3-7.0 (green). Predicted GMST (dashed) is compared with NorESM2 (solid), with shading indicating $\pm2$ standard deviations across 50 trajectories. Our model is able to distinguishes the scenario-dependent warming trends and captures the stronger warming under the held-out SSP3-7.0 scenario.
    (b) CO$_2$, spatially averaged CH$_4$, BC, and spatially averaged SO$_2$
    forcings are shown for SSP1-2.6, SSP2-4.5, and SSP3-7.0 over the 85-year
    evaluation period. SSP1-2.6 and SSP2-4.5 represent lower- and intermediate-forcing pathways, respectively, while SSP3-7.0 represents a scenario with higher greenhouse gas and aerosol concentrations.
    }
    \label{fig:3ssp}
\end{figure}

\begin{figure}[htbp!]
    \centering
    \includegraphics[width=.65\textwidth]{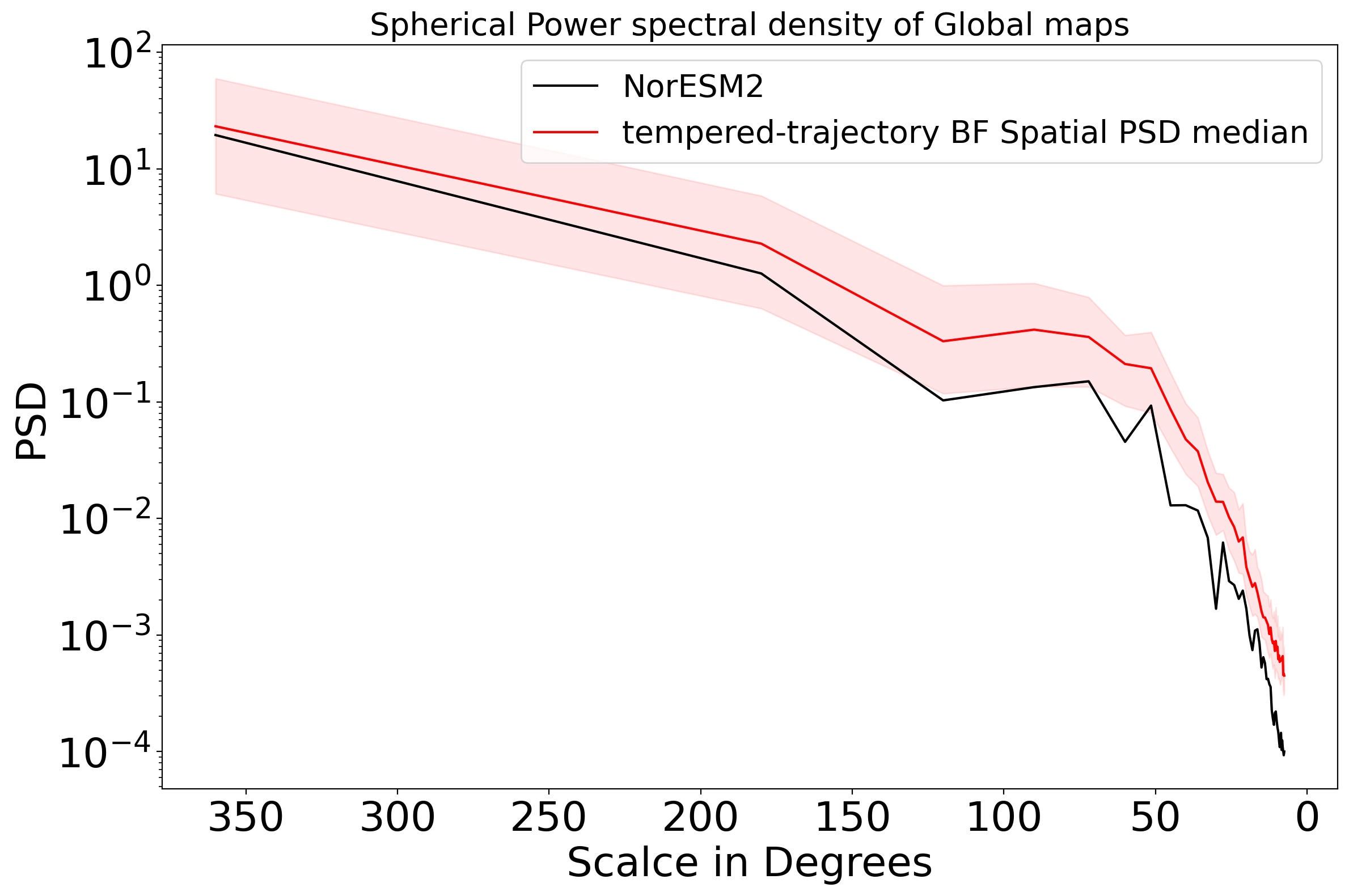}
    \caption{\textbf{Spherical power spectral density (PSD) of the time-averaged global surface temperature field.} For each trajectory, monthly surface temperature maps are first averaged over time, followed by a spherical harmonic decomposition of the resulting global HEALPix map. The spatial power spectrum is represented by the angular power spectrum $C_{\ell}$. The horizontal axis shows the approximate angular scale, $360^\circ/(\ell+1)$, where smaller values correspond to finer spatial structures. The black curve denotes the NorESM2 reference, while the red curve shows the median PSD across model trajectories, with the shaded region indicating the 16th--84th percentile range. Both spectra are dominated by large-scale spatial structures, with power decreasing toward finer scales. The model closely reproduces the overall spectral slope and scale hierarchy of NorESM2, while it overestimates the power at fine spatial scales.}
    \label{fig:psd_spatial}
\end{figure}

\begin{figure}[htbp!]
    \centering
    \includegraphics[width=\textwidth,trim={0cm 7.5cm 0cm 0cm},clip]{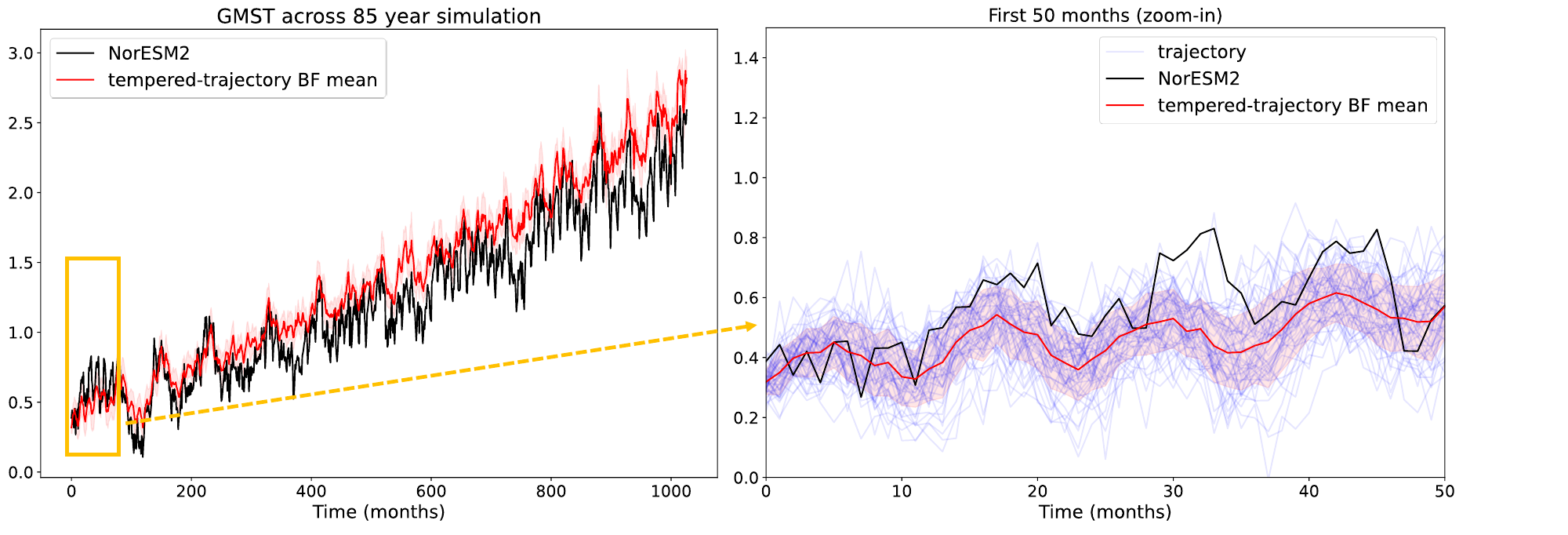}
    \caption{\textbf{GMST across 85 year rollouts  under SSP3-7.0.} The left panel shows the global mean surface temperature (GMST) over the full 85-year autoregressive rollout. The right panel provides a detailed view of the first 50 months, showing the ensemble mean and standard deviation across 50 trajectories, together with all 50 individual sample trajectories.}
    \label{fig:gmst_ssp3}
\end{figure}

\begin{figure}[htbp!]
    \centering
    \includegraphics[width=\textwidth,trim={0cm 7.5cm 0cm 0cm},clip]{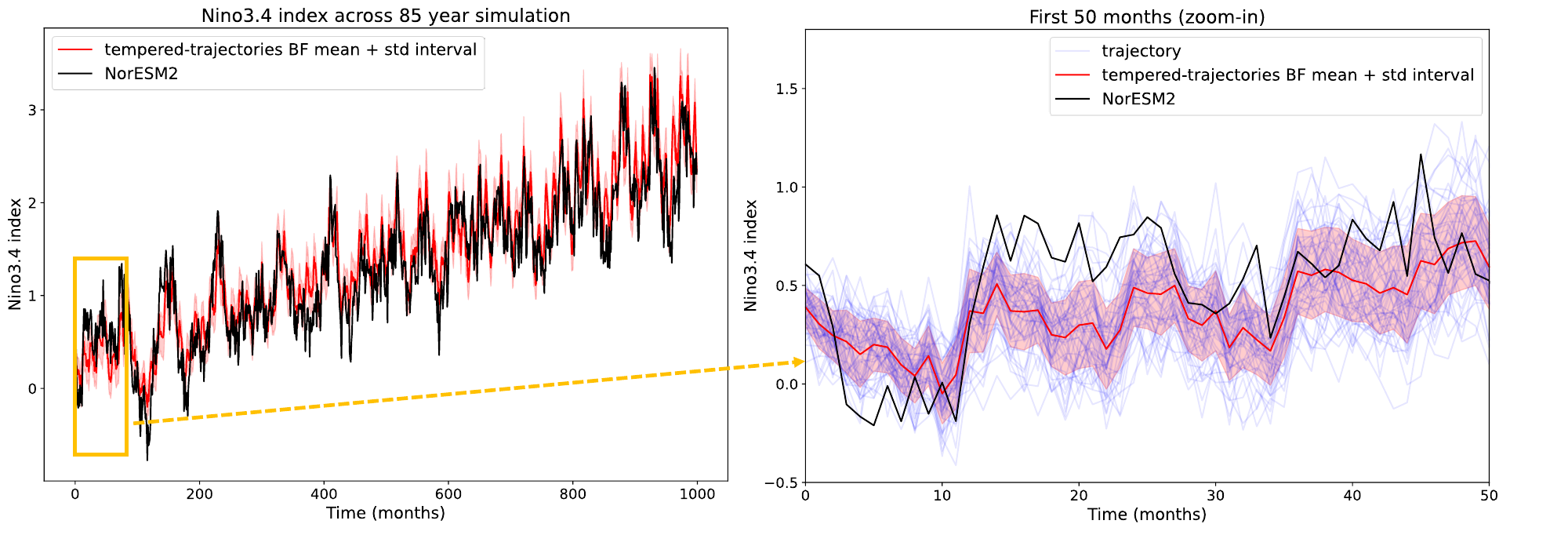}
    \caption{\textbf{Ni\~no3.4 index across 85-year rollouts under SSP3-7.0.}
    The left panel shows the Ni\~no3.4 index over the full 85-year autoregressive rollout. 
    The right panel provides a detailed view of the first 50 months, showing the mean and standard deviation of 50 trajectories, and 50 individual sample trajectories.}
    \label{fig:nino_ssp3}
\end{figure}

\begin{figure}[htbp!]
    \centering
    \includegraphics[width=\textwidth,trim={0cm 7.5cm 0cm 0cm},clip]{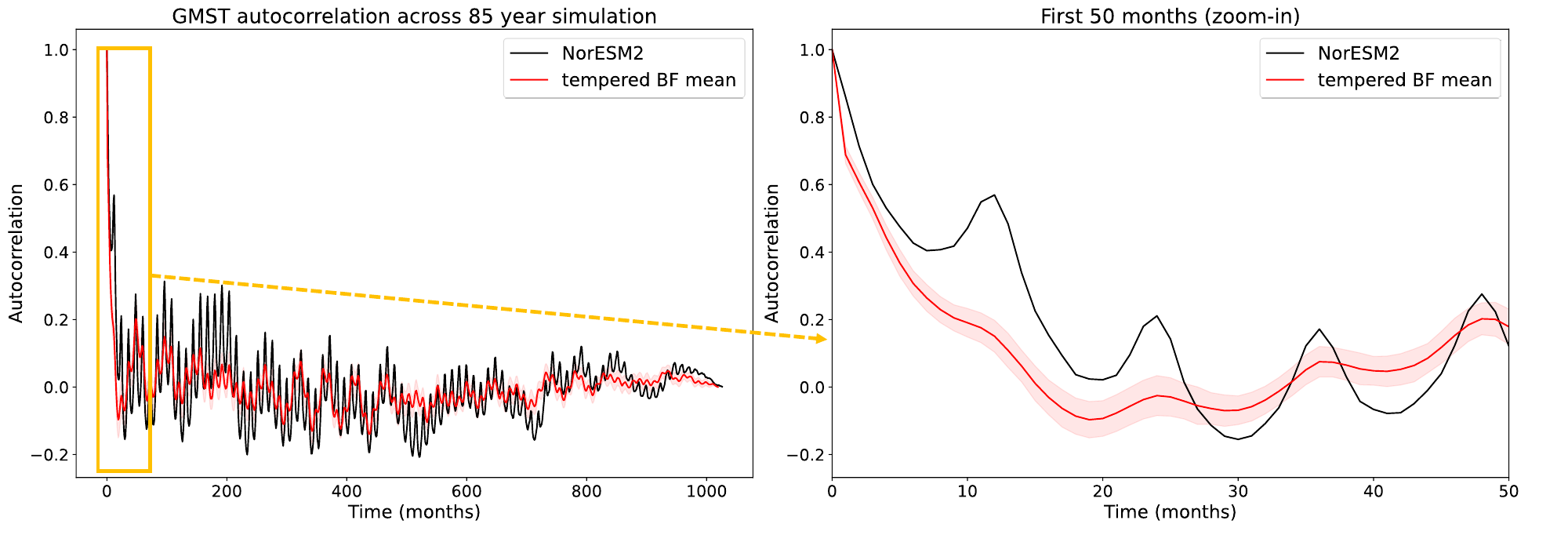}
\caption{\textbf{GMST time-series autocorrelation under SSP3-7.0.} The left panel shows the autocorrelation of GMST over the full 85-year autoregressive rollout, comparing the model prediction with the NorESM2 ground truth. The right panel provides a detailed view of the first 50 months, highlighting the short-term temporal dependence and seasonal oscillations. Both time series are linearly detrended prior to computing the autocorrelation to remove the long-term warming trend and prevent it from dominating the autocorrelation at large time lags. The model captures the approximately annual oscillatory structure of the reference autocorrelation, while exhibiting weaker short-term persistence.}
    \label{fig:autocorrelation_gmst}
\end{figure}

\begin{figure}[htbp!]
    \centering
    \includegraphics[width=\textwidth,trim={0cm 7.5cm 0cm 0cm},clip]{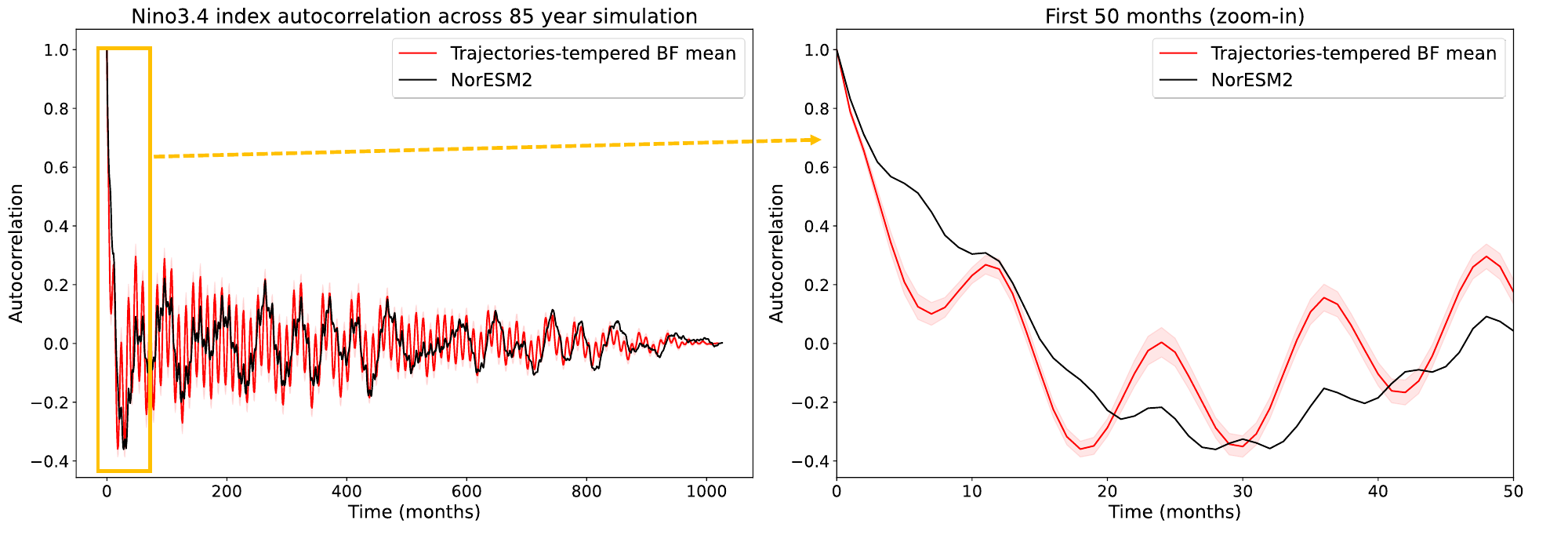}
    \caption{\textbf{Ni\~no3.4 time-series autocorrelation under SSP3-7.0.}
The left panel shows the autocorrelation of the Ni\~no3.4 index over the full 85-year autoregressive rollout, comparing the model prediction with the NorESM2 ground truth.
The right panel provides a detailed view of the first 50 months, highlighting the short-term temporal dependence and oscillatory variability.
Both time series are linearly detrended prior to computing the autocorrelation to remove the long-term trend and prevent it from dominating the autocorrelation at large time lags. Compared with GMST, the model more closely reproduces the temporal dependence of the Ni\~no3.4 index at one-year lag.}
    \label{fig:autocorrelation_nino}
\end{figure}

\begin{figure}[htbp!]
    \centering
    \includegraphics[width=\textwidth,trim={0cm 7.5cm 0cm 0cm},clip]{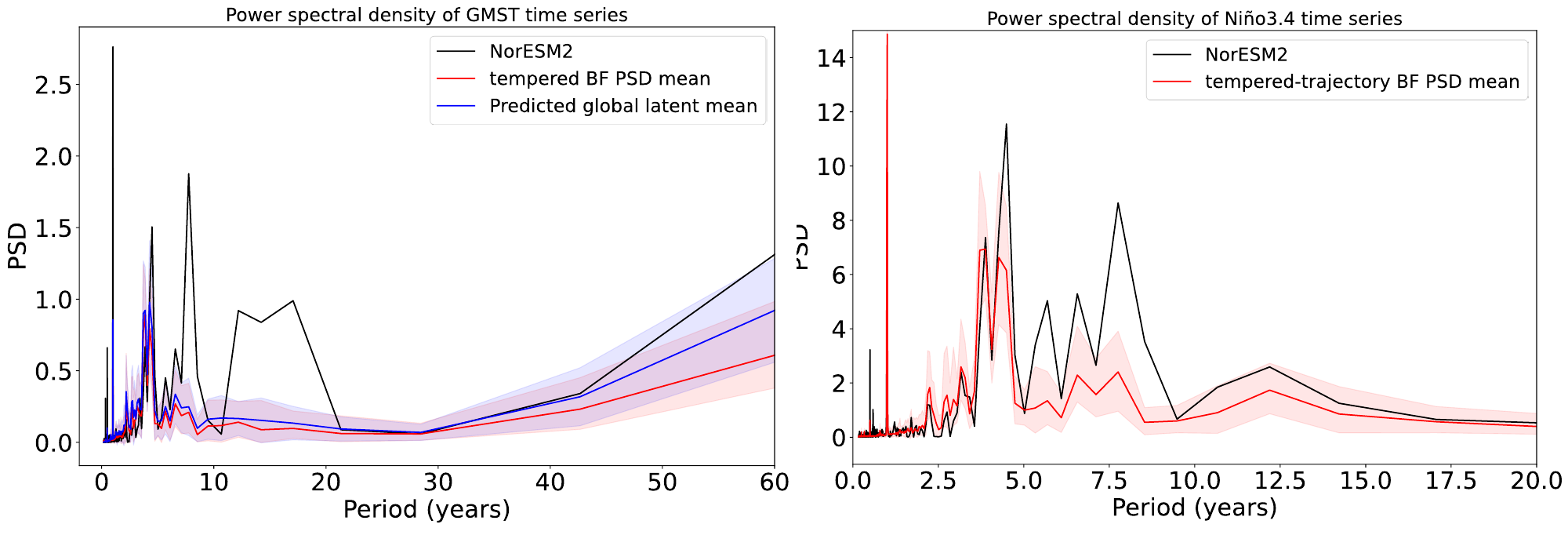}
\caption{\textbf{Power spectral density (PSD) of the GMST and Ni\~no3.4 index time series under SSP3-7.0.}
The left panel shows the PSD of the GMST time series, while the right panel shows the PSD of the Ni\~no3.4 index.
The model PSD is computed across 50 autoregressive rollout trajectories, with the solid red line indicating the ensemble mean and the shaded region indicating $\pm 1$ standard deviation.
The black line denotes the NorESM2 ground-truth spectrum.
In the GMST panel, the blue line additionally shows the PSD of the learned global climate latent, illustrating its temporal variability across different time scales.
All time series are linearly detrended prior to PSD estimation. Global latent, compared with the spatial mean of the predicted temperature field, is better at capturing the long-term variability of GMST.
The model reproduces the major spectral peaks of the Ni\~no3.4 index, including the annual peak and variability at approximately 2 to 5 year periods.}
    \label{fig:pds_time_series}
\end{figure}

We provide additional analyses of the long-term rollouts, including the SSP forcings (Fig.~\ref{fig:3ssp_forcings}) and the spatial PSD of the time-averaged temperature fields (Fig.~\ref{fig:psd_spatial}). We further report the GMST (Fig.~\ref{fig:gmst_ssp3}) and Ni\~no3.4 time series (Fig.~\ref{fig:nino_ssp3}), together with their temporal autocorrelations  (Fig.~\ref{fig:autocorrelation_gmst} and Fig.~\ref{fig:autocorrelation_nino}) and PSDs (Fig.~\ref{fig:pds_time_series}). GMST is computed by spatially averaging the temperature field, while Ni\~no3.4 is an El Ni\~no-Southern Oscillation (ENSO) index computed by averaging sea-surface temperature anomalies over the Ni\~no3.4 region in the equatorial Pacific.

\subsection{Baselines and Ablations}\label{app:ablation}
We compare against the linear pattern scaling (LPS) baseline~\citep{bjorn_linear}, which first regressing GMST on GHG, and then learning a point-wise linear mapping from GMST to the spatial temperature field. Despite its simplicity, LPS has been shown to outperform several deep-learning climate emulators. Our model better captures the mean GMST and Ni\~no3.4 statistics, while LPS more accurately reproduces the GMST range. However, LPS substantially underestimates the Ni\~no3.4 range and captures primarily the long-term trend. Consequently, it fails to reproduce the characteristic spectral variability of Ni\~no3.4 time series, resulting in a near-zero PSD at interannual timescales (Fig.~\ref{fig:result}b) after detrending.
 
\begin{figure}[htbp!]
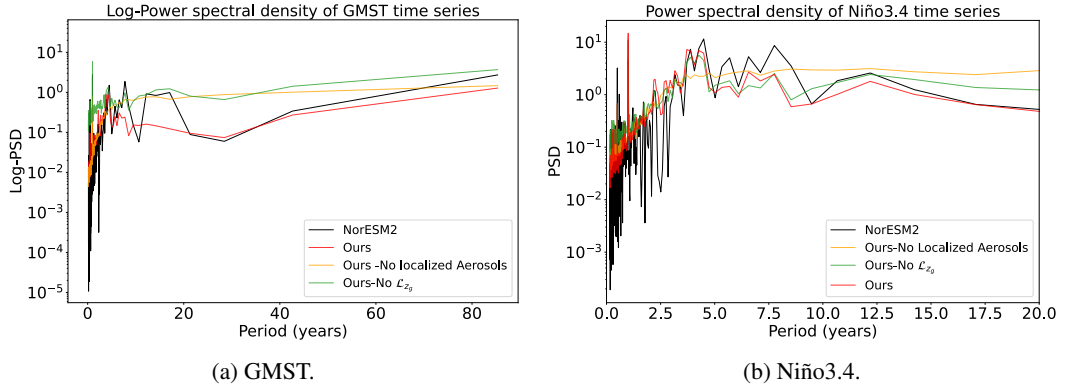

    \centering
    \begin{subfigure}[t]{0.49\linewidth}
        \centering
        \includesvg[width=\linewidth]{Figures/Figure_pz2mu/gmst_psd_time_log_ablation}
        \caption{GMST.}
        \label{fig:gmst_psd_ablation}
    \end{subfigure}
    \hfill
    \begin{subfigure}[t]{0.49\linewidth}
        \centering
        \includesvg[width=\linewidth]{Figures/Figure_pz2mu/Nino_psd_time_log_ablation}
        \caption{Ni\~no3.4.}
        \label{fig:nino_psd_ablation}
    \end{subfigure}

    \caption{\textbf{Ablation study of the temporal power spectral density (PSD) under SSP3-7.0, shown on a logarithmic scale.}
    \textbf{(a)} GMST and \textbf{(b)} Ni\~no3.4 PSDs from 85-year autoregressive rollouts. We compare the full model with variants without the local aerosol mapping and without the global-latent alignment loss $\mathcal{L}_{z_g}$. Removing $\mathcal{L}_{z_g}$ amplifies temporal variability, especially over the low-frequency part. Removing the local mapping produces a smoother PSD and fails to capture the characteristic Ni\~no3.4 peak.}
    \label{fig:psd_log_ablation}
\end{figure}

We also compare the full model with two ablations. \textit{No Localized Aerosols} removes the localized aerosol assignment, allowing aerosol forcings from all locations to affect each local latent. \textit{No $\mathcal{L}{z_g}$} removes the alignment between the global latent and GMST. Without this alignment, its rollouts follow the particle selection procedure in~\cite{picabu}. Table~\ref{tab:statistic_score} summarizes the rollout statistics, while Fig.~\ref{fig:psd_log_ablation} compares the temporal PSDs of GMST and Ni\~no3.4. Removing the localized mapping improves the statistics for both indices, but suppresses temporal variability, producing an overly smooth Ni\~no3.4 PSD and failing to capture its characteristic peak. Removing $\mathcal{L}{z_g}$ degrades the statistical scores and leads to larger long-term drift from the NorESM2 reference.

\begin{table}[htbp!]
    \centering
        \caption{\textbf{Statistics of long-term climate rollouts under SSP3-7.0 forcing}. Starting from the climate state on 2015-01-01, each model is rolled out autoregressively for 85 years, conditioned at each time step on the corresponding SSP3-7.0 external forcings. The particle selection follows~\ref{sec:method_rollouts}. The left section shows the mean, standard deviation, range of the GMST, and the same for the Niño3.4 index on the right. The first row shows the reference statistics computed from the ground-truth NorESM2 SSP3-7.0 simulation, followed by comparison method, and then ablations. For all metrics, being closer to the ground truth is better.}
    \resizebox{\textwidth}{!}{%
    \begin{tabular}{cl|ccc|ccc}
    \toprule
    & ~ & \multicolumn{3}{c|}{\textbf{GMST}} & \multicolumn{3}{c}{\textbf{Niño3.4}} \\
    \cmidrule(lr){3-5}
    \cmidrule(lr){6-8}
    
    & ~ & \textbf{Mean} & \textbf{Std. Dev.} & \textbf{Range} & \textbf{Mean} & \textbf{Std. Dev.} & \textbf{Range}\\\midrule
    
    & Ground truth & 1.2403 & 0.5771  &  2.5144 &  1.3662 & 0.7751  &  4.2305\\\midrule
    
    & Linear Pattern Scaling~\cite{bjorn_linear} & 1.4711  & 0.7009 & \textbf{2.4247} & 1.5677 & \textbf{0.7383} & 2.5535  \\
    % & Ours-No Localized Aerosols & 0.9838 & 0.6637 & 5.0812 & 0.9046 & 0.6012 & \textbf{4.6899} \\
    & Ours-No Localized Aerosols & \textbf{1.3281} & \textbf{0.5327} & 2.7091 & \textbf{1.4013} & 0.6702 & \textbf{4.3710} \\ %-> filtering using global latent
    & Ours-No $\mathcal{L}_{z_g}$ & 1.0015 & 0.6247 & 4.8010 & 1.1235 & 0.6663 & 4.7029 \\
    % & Ours-No $\mathcal{L}_{z_g}$ & -- &-- & -- & 1.7780 & 0.6159 & 4.3608 \\%-> filtering using global latent
    % & Ours &  \textbf{1.3322}  & 0.6607 & 4.3800 & \textbf{1.4045}  & \textbf{0.7744} & 5.0484 \\
    & Ours & 1.4107 & 0.6573 & 3.2710 &  1.4692 & 0.8205 & 4.7190\\ %-> filtering using global latent
    \bottomrule
    \end{tabular}%
    }
    \label{tab:statistic_score}
\end{table}

\subsection{Additional counterfactual experiments}\label{app:counterfactual}
\begin{figure}[t]
    \centering
    \begin{subfigure}[t]{0.85\linewidth}
        \centering
        \includegraphics[width=\linewidth,trim={0cm 0cm 0cm 2.5cm},clip]{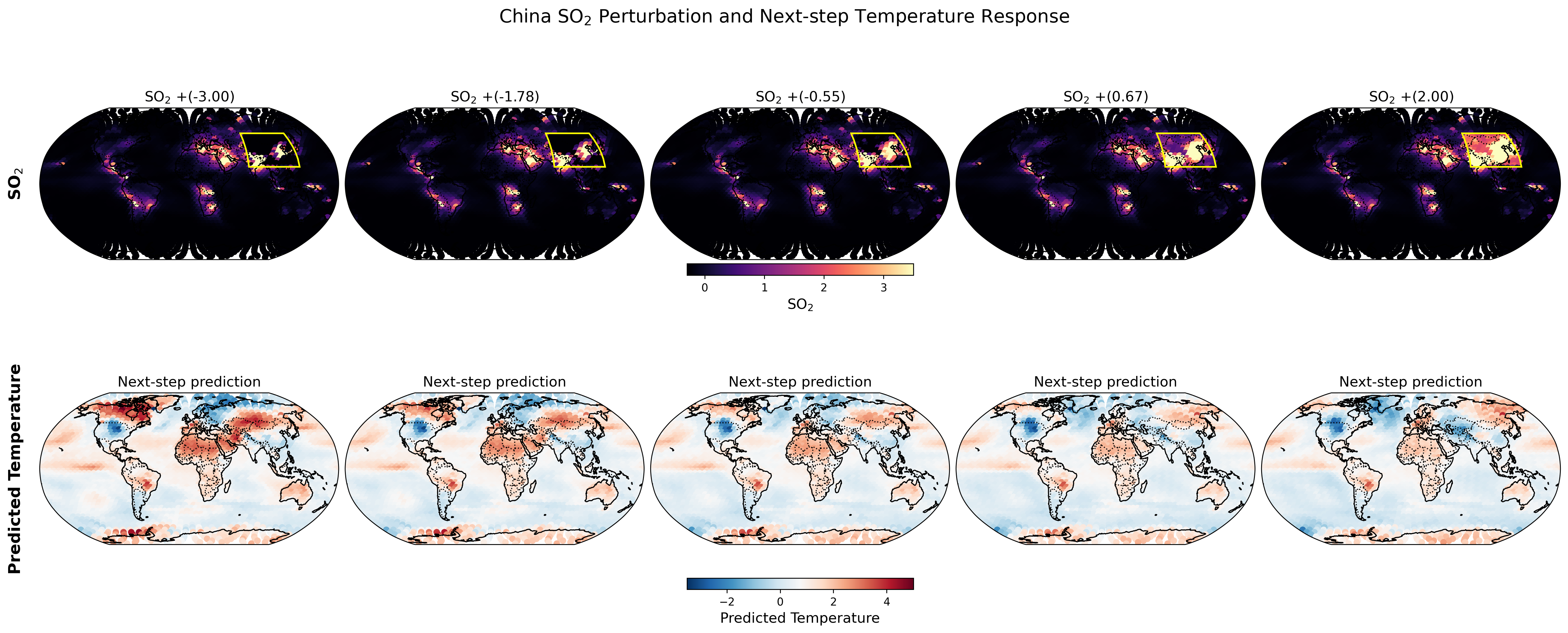}
        \caption{ Regional SO$_2$ intervention. }
        \label{fig:perturbation_so2_samples}
    \end{subfigure}
    \hfill
    \begin{subfigure}[t]{0.85\linewidth}
        \centering
        \includegraphics[width=\linewidth,trim={0cm 0cm 0cm 2.5cm},clip]{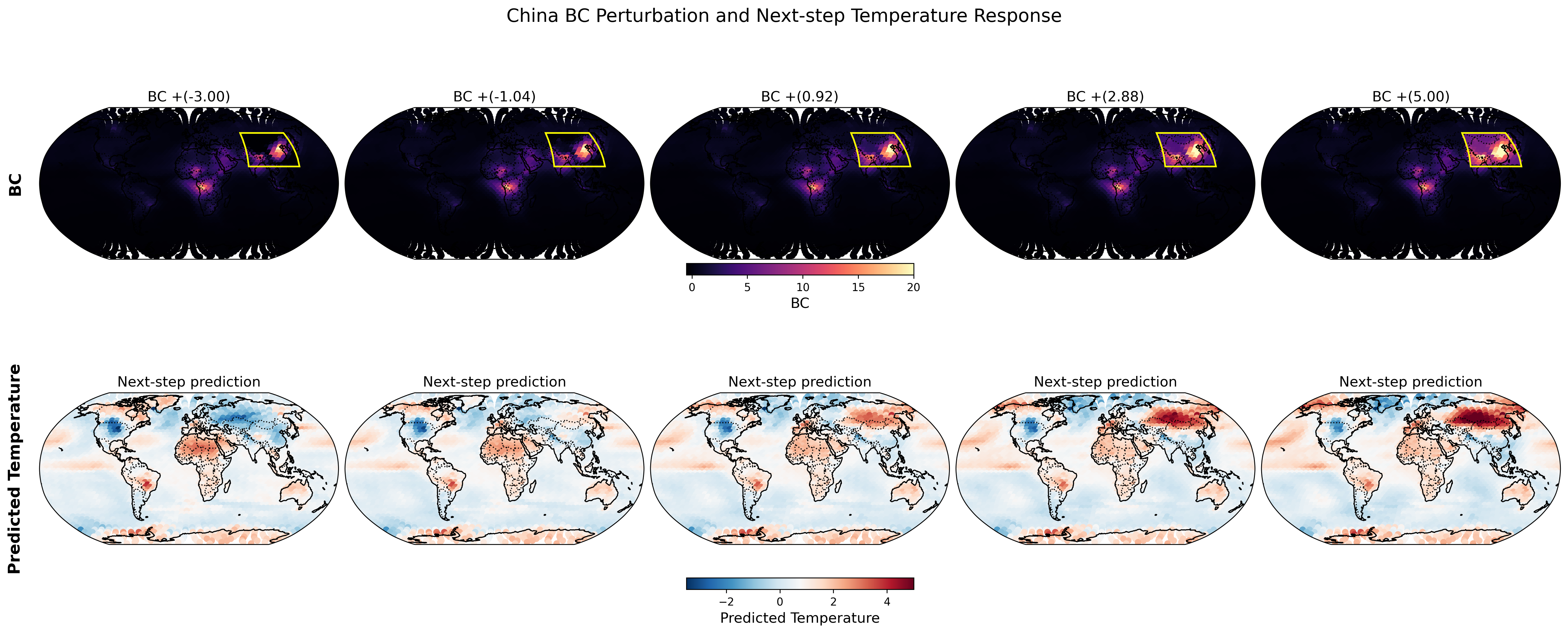}
        \caption{Regional BC perturbations. }
        \label{fig:perturbation_bc_samples}
    \end{subfigure}

    \caption{\textbf{Sample visualization of the temperature response to aerosol perturbations.} The upper row shows five different SO$_2$ perturbation levels applied within East Asia, and the lower row shows the corresponding next-step temperature predictions. Increasing SO$_2$ produces a cooling response, whereas increasing BC produces a warming response, with the strongest changes concentrated around the perturbed region.}
    \label{fig:perturbation_aerosols_samples}
\end{figure}

\begin{figure}[htbp!]
    \centering
    \includegraphics[width=1\textwidth]{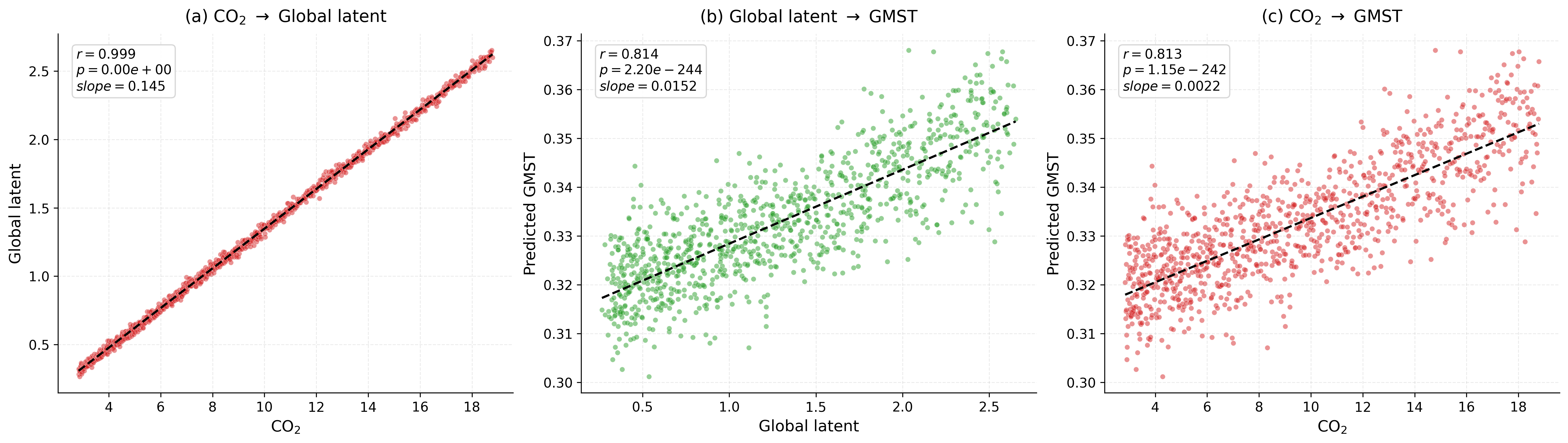}
\caption{\textbf{Single-step response to CO$_2$ perturbations.}
We evaluate the model under controlled CO$_2$ perturbations while keeping the input climate state and all other forcing variables fixed. Starting from a random selected initial condition, CO$_2$ is varied across its range of values observed under SSP3-7.0, and a single-step prediction is performed for each forcing value. (a) Relationship between the perturbed CO$_2$ concentration and the predicted global climate latent ($\mu_{z_{g}}$). (b) Relationship between the predicted global latent and the corresponding predicted GMST. (c) Direct relationship between the perturbed CO$_2$ concentration and predicted GMST. Dashed lines indicate linear fits, with the Pearson correlation coefficient ($r$), $p$-value, and fitted slope reported in each panel.
The results show a nearly linear response of the global latent to CO$_2$ perturbations, which is subsequently reflected in the predicted GMST.}
    \label{fig:scatter_response_co2}
\end{figure}

\begin{figure}[htbp!]
    \centering
    \includegraphics[width=1\textwidth]{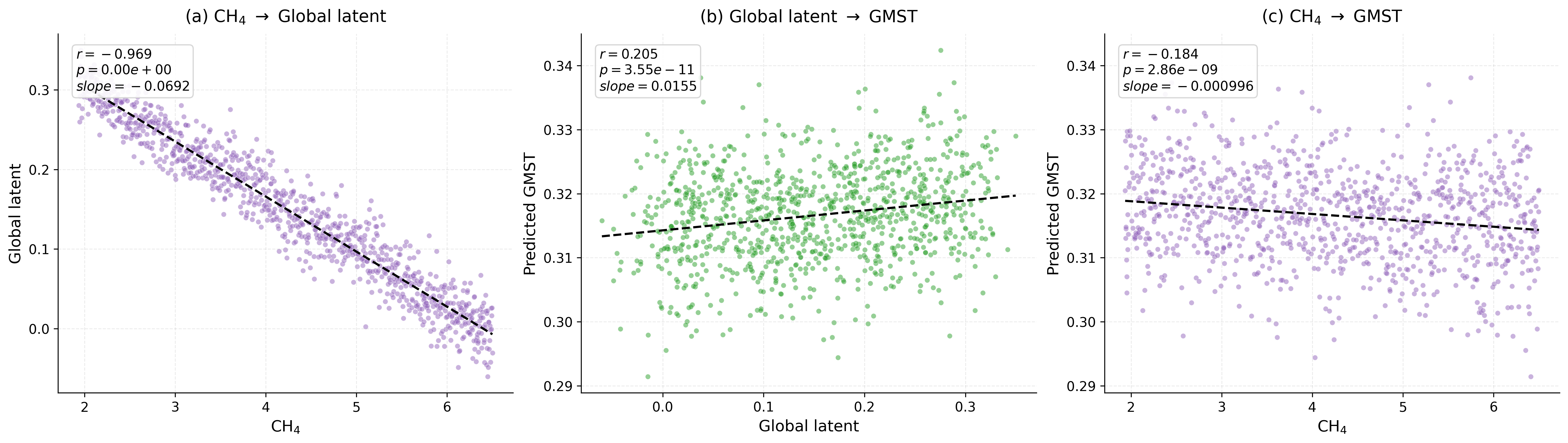}
    \caption{\textbf{Single-step response to CH$_4$ perturbations.} Compared with CO$_2$, CH$_4$ exhibits a weaker modulation of the learned global latent, and this variation is propagated even more weakly to the next-step GMST prediction, resulting in a substantially weaker direct CH$_4$--GMST response. The cooling response to increased CH$_4$ reflects a limitation of the current model in disentangling the effects of strongly correlated global forcings and remains an area for further improvement.}
    \label{fig:scatter_response_ch4}
\end{figure}

\begin{figure}[htbp!]
    \centering
    \includegraphics[width=1\textwidth]{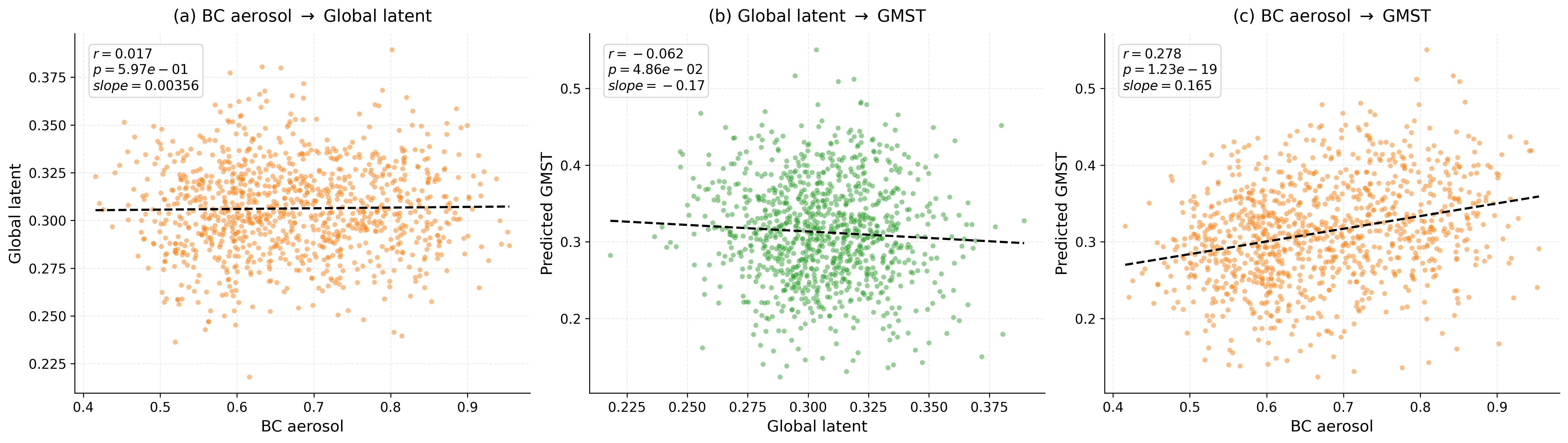}
    \caption{\textbf{Single-step response to BC perturbations.} In contrast to the global greenhouse-gas forcings, BC shows almost no modulation of the learned global latent ($r=0.017$), while exhibiting a clearer positive response in the predicted GMST ($r=0.278$). This suggests that the modeled BC temperature response is primarily mediated through the local climate latents rather than the global latent.}
    \label{fig:scatter_response_aerosol}
\end{figure}

\begin{figure}[htbp!]
    \centering
    \includegraphics[width=1\textwidth]{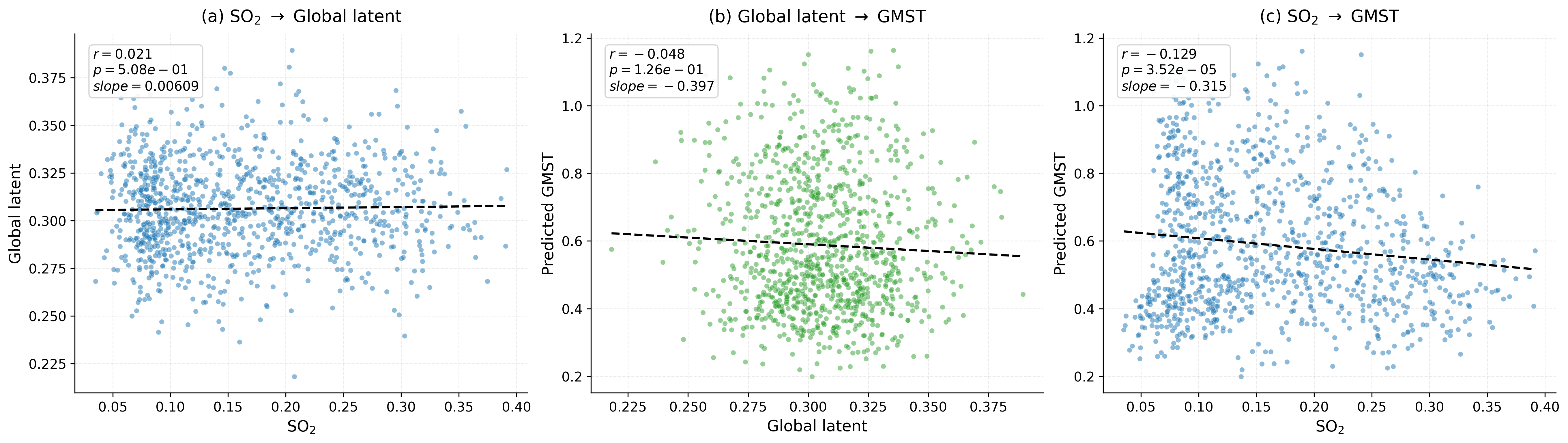}
    \caption{\textbf{Single-step response to SO$_2$ perturbations.} Similar to BC, SO$_2$ shows almost no modulation of the learned global latent ($r=0.021$), while exhibiting a weak negative response in the predicted GMST ($r=-0.129$). The global latent responds to global forcing perturbations but remains insensitive to aerosol perturbations, providing evidence that the model separates global and localized forcings through distinct causal pathways.}
    \label{fig:scatter_response_so2}
\end{figure}

\begin{figure}[htbp!]
    \centering
    \includegraphics[width=1\textwidth,trim={0cm 0cm 0cm 2.5cm},clip]{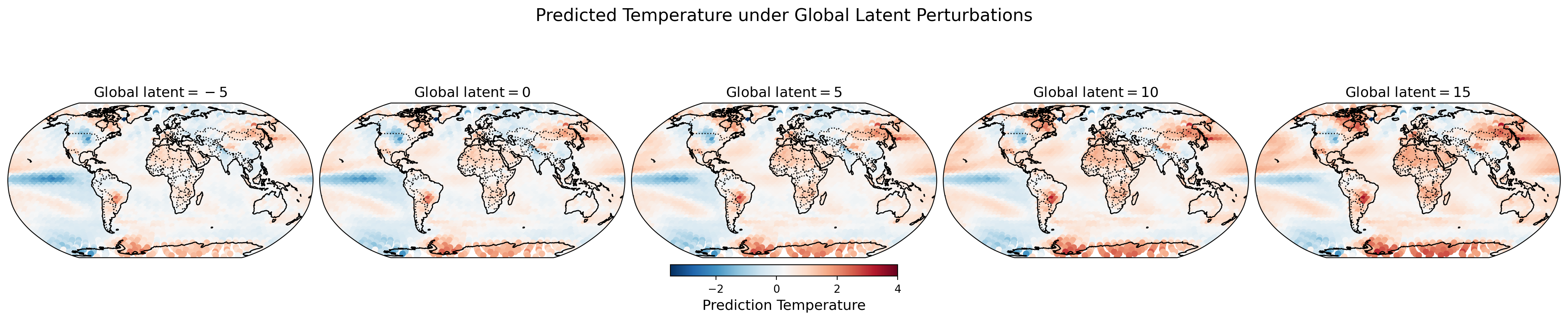}
    \caption{\textbf{Temperature response to interventions on the predicted global climate latent.}
We directly intervene on the predicted global latent mean ($\mu_{z_{g}}$) by setting it to $-5$, $0$, $5$, $10$, and $15$, while keeping the input climate history and external forcing conditions unchanged. For each intervention, the prescribed global latent is passed to the local climate transition model, and the resulting local latent state is decoded into the predicted surface temperature field.
Each panel shows the prediction for the same input sample under a different value of the global latent. As the global latent increases, the predicted temperature field exhibits a progressively stronger warming response over much of the globe.
This sensitivity experiment indicates that the learned global latent modulates the large-scale temperature response.}
    \label{fig:perturb_pz2mu}
\end{figure}

We provide additional analyses of the forcing-perturbation experiments, including sampled temperature responses to regional aerosol perturbations (Fig.~\ref{fig:perturbation_aerosols_samples}), single-step forcing--response pathways for each forcing variable (Fig.~\ref{fig:scatter_response_co2} -- Fig.~\ref{fig:scatter_response_so2}), and the GMST response to perturbations of the learned global latent $z_g$ (Fig.~\ref{fig:perturb_pz2mu}). Together, these analyses further characterize how global and regional forcing signals propagate through the learned representation to the predicted climate state.

\begin{figure}
    \centering
    \includesvg[width=.5\linewidth]{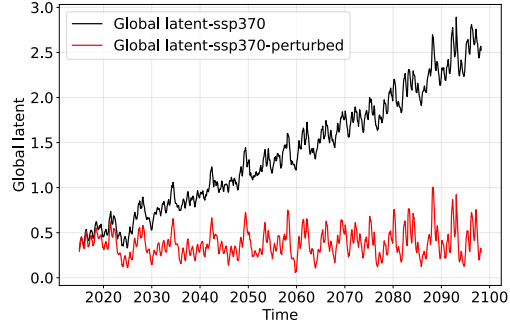}
    \caption{\textbf{Global latent response to forcing perturbations in long-term rollout.}
Global latent trajectories under SSP3-7.0 with original forcings (black) and with CO$_2$ and CH$_4$ fixed at their 2015 levels (red), showing a substantially suppressed long-term increase under the perturbation.}
    \label{fig:rollout_pz2mu}
\end{figure}

